\documentclass[letterpaper]{article} 
\usepackage{aaai2026}  
\usepackage{times}  
\usepackage{helvet}  
\usepackage{courier}  
\usepackage[hyphens]{url}  
\usepackage{graphicx} 
\usepackage{natbib}  
\usepackage{caption} 
\usepackage{algorithm}
\usepackage{algorithmic}
\usepackage{amsmath}

\usepackage{tcolorbox}

\usepackage{newfloat}
\usepackage{listings}
\DeclareCaptionStyle{ruled}{labelfont=normalfont,labelsep=colon,strut=off} 
\floatstyle{ruled}
\newfloat{listing}{tb}{lst}{}
\floatname{listing}{Listing}
\title{AgroAskAI: A Multi-Agentic AI Framework for Supporting Smallholder Farmers' Enquiries Globally}
\author {
    Nadine Angela Cantonjos,
    Arpita Biswas
}
\affiliations {
    Rutgers University\\
    npc58@scarletmail.rutgers.edu, a.biswas@rutgers.edu
}

\usepackage{bibentry}

\begin{document}

\maketitle

\begin{abstract}
Agricultural regions in rural areas face damage from climate-related risks, including droughts, heavy rainfall, and shifting weather patterns. Prior research calls for adaptive risk-management solutions and decision-making strategies. To this end, artificial intelligence (AI), particularly agentic AI, offers a promising path forward. Agentic AI systems consist of autonomous, specialized agents capable of solving complex, dynamic tasks. While past systems have relied on single-agent models or have used multi-agent frameworks only for static functions, there is a growing need for architectures that support dynamic collaborative reasoning and context-aware outputs. To bridge this gap, we present AgroAskAI, a multi-agent reasoning system for climate adaptation decision support in agriculture, with a focus on vulnerable rural communities. AgroAskAI features a modular, role-specialized architecture that uses a chain-of-responsibility approach to coordinate autonomous agents, integrating real-time tools and datasets. The system has built-in governance mechanisms that mitigate hallucination and enable internal feedback for coherent, locally relevant strategies. The system also supports multilingual interactions, making it accessible to non-English-speaking farmers. Experiments on common agricultural queries related to climate adaptation show that, with additional tools and prompt refinement, AgroAskAI delivers more actionable, grounded, and inclusive outputs. Our experimental results highlight the potential of agentic AI for sustainable and accountable decision support in climate adaptation for agriculture.

\end{abstract}

\begin{links}
    \link{Code}{https://github.com/ncantonjos04/AgroAskAI}
\end{links}


\section{Introduction}
As agriculture faces a growing strain from the environmental and socioeconomic disruptions of climate change, artificial intelligence (AI) enables new ways of providing dynamic decision support to farming communities. In these communities, agriculture is not only a primary livelihood but also a fundamental component of local economies and daily life. However, challenges posed by climate-related risks, such as erratic weather, depleted soil health, and shifted rainfall, have negatively affected farmers in vulnerable communities. For instance, crop production in rural regions of Central America has declined over the past two decades due to more frequent storms and severe flooding~\cite{waddick2017effects}. In the arid and semi-arid regions of Kenya, $31\%$ of smallholder farmers lack access to climate-related information, leaving them unprepared to adapt to rising temperatures, prolonged droughts, and increasingly unpredictable rainfall patterns~\cite{mutunga2017smallholder}. 

The severity of these local agricultural challenges is amplified by global environmental degradation and economic instability. Extreme weather-related events caused nearly $\$1.5$ trillion in economic loss worldwide in the decade to 2019, a substantial increase from $\$184$ billion in the 1970s~\cite{charlton2023climate,wangfuture}. In 2023, droughts in the Southern and Midwestern regions had caused $\$14.5$ billion loss, impacting the agriculture sector, including damage to field crops from lack of rainfall and heat~\cite{smith2023historic,silici2021adaptive}.  
These trends underscore the need for improved risk management and decision support to prevent a vast amount of damage from occurring in the future. 

Previous studies have examined climate adaptation decisions among smallholder farmers in specific regions, such as in the mountainous areas of Vietnam~\cite{hoa2022exploring}, far-western Nepal~\cite{lamichhane2022nepal}, and the semi-arid zones of Senegal~\cite{zagre2024senegal}. While these localized insights are valuable, increasing global variability in climate patterns and the rising frequency of extreme weather events underscore the urgent need for effective, scalable adaptation strategies. In particular, smallholder farmers across the globe, who are among the most vulnerable to climate disruptions, require decision support to navigate these complex and evolving challenges. 

Effective adaptation measures often include the development of climate-resilient crop varieties, improved cultivation practices, and timely dissemination of climate information~\cite{castellanos2019climate}. However, the successful scaling of such innovative practices across diverse agricultural landscapes demands solutions that are both context-specific and locally adaptable. One-size-fits-all solutions risk not only ineffectiveness but may also lead to maladaptation, compounding existing agricultural vulnerabilities. Therefore, systems that support climate resilience must be not only intelligent but also dynamic—capable of adjusting their recommendations and suggestions based on real-time data and feedback, in order to advance climate resilience among smallholder farmers.


This is where agentic AI offers transformative potential. Unlike conventional AI systems that require frequent human oversight, agentic AI systems exhibit a higher degree of autonomy. They can reason through complex, multi-step objectives, integrate historical and real-time external data, and independently execute decision processes from end to end~\cite{mohammedagentic}. Applied to agriculture, agentic AI can generate novel, tailored, and context-aware solutions that respond adaptively to changing environmental and socio-economic conditions. 
Recent work by \citet{raza2025trism} highlights the importance of integrating Trust, Risk, and Security Management (TRiSM) into agentic AI systems, emphasizing transparent reasoning, policy-based oversight, and regulatory traceability as key pillars for ethical and reliable AI deployment. However, as they note, no unified framework currently operationalizes these principles within real-world, autonomous AI systems—especially in high-stakes domains like agriculture.

In this paper, we introduce \textbf{AgroAskAI}, a modular, agentic, multi-agent AI framework designed to support smallholder farmers by enhancing agricultural decision-making under uncertainty. AgroAskAI integrates role-specialized autonomous agents, real-time external resources (e.g., weather data, forecast model, geospatial information), and a critique mechanism into a multi-agent architecture in order to ensure traceability and context-aware reasoning. In addition, AgroAskAI implements explicit reasoning logs, decision checkpoints, and external validation to promote auditability. A dedicated critique (reviewer agent) embeds correctness and feasibility as first-class objectives to reduce hallucination while ensuring locally grounded, actionable plans. In addition to decision support, AgroAskAI plays an educational role by making its reasoning process transparent, fostering trust and understanding. By autonomously generating adaptive, context-specific strategies, AgroAskAI equips farmers with the tools and knowledge to build lasting agricultural resilience in the face of climate change.

\section{Related Work}
Artificial Intelligence (AI) is increasingly being leveraged to support farmers in managing the complex challenges posed by climate change in agriculture. Prior work has explored AI-driven systems for automated sensor monitoring of soil and crop health~\cite{farmbeats}, often aimed at optimizing yield. While such systems are valuable, they fall outside the scope of this work. In contrast, our focus is on developing a conversational AI system designed to support farmers through natural language interactions, rather than through sensor-based automation.

The Agrifriendly Conversational AI Chatbot~\cite{gujjar2025agri} uses a single agent, generative system based on the Rasa framework (https://rasa.com/) to understand natural human language and assist farmers with queries related to crop maintenance. While effective for basic queries, it relies on a single-agent model that limits adaptability and lacks internal critique or modular reasoning. Farmer.Chat~\cite{singh2024farmer} introduces a generative chatbot deployed via platforms like WhatsApp and Facebook Messenger, using Retrieval-Augmented Generation (RAG) to deliver informative, multilingual responses. Its governance relies on user feedback loops and data shared through local partners. AgriBuddy~\cite{tonmoyagribuddy} is an agentic chatbot system, tailored for Bangladeshi smallholder farmers supporting interaction in Bangla, English, and hybrid Bangla-English. The system is composed of specialized agents, such as a query handler, memory agent, and expert advisory agent, and includes a CNN (Convolutional Neural Network) vision model for rice disease detection. 

In contrast to prior research, AgroAskAI provides reliable climate-adaptive solutions to smallholder farmers across the globe by adopting a modular, multi-agent architecture with embedded internal governance mechanisms that proactively mitigate hallucination and ensure output reliability. Instead of relying on external approvals, the system accesses reliable real-time information from trusted historical and forecasted weather data, while also allowing users to contribute verified content through a curated external document repository, enhancing both transparency and adaptability.

\begin{figure*}[!h]
    \centering
    \includegraphics[scale=0.53]{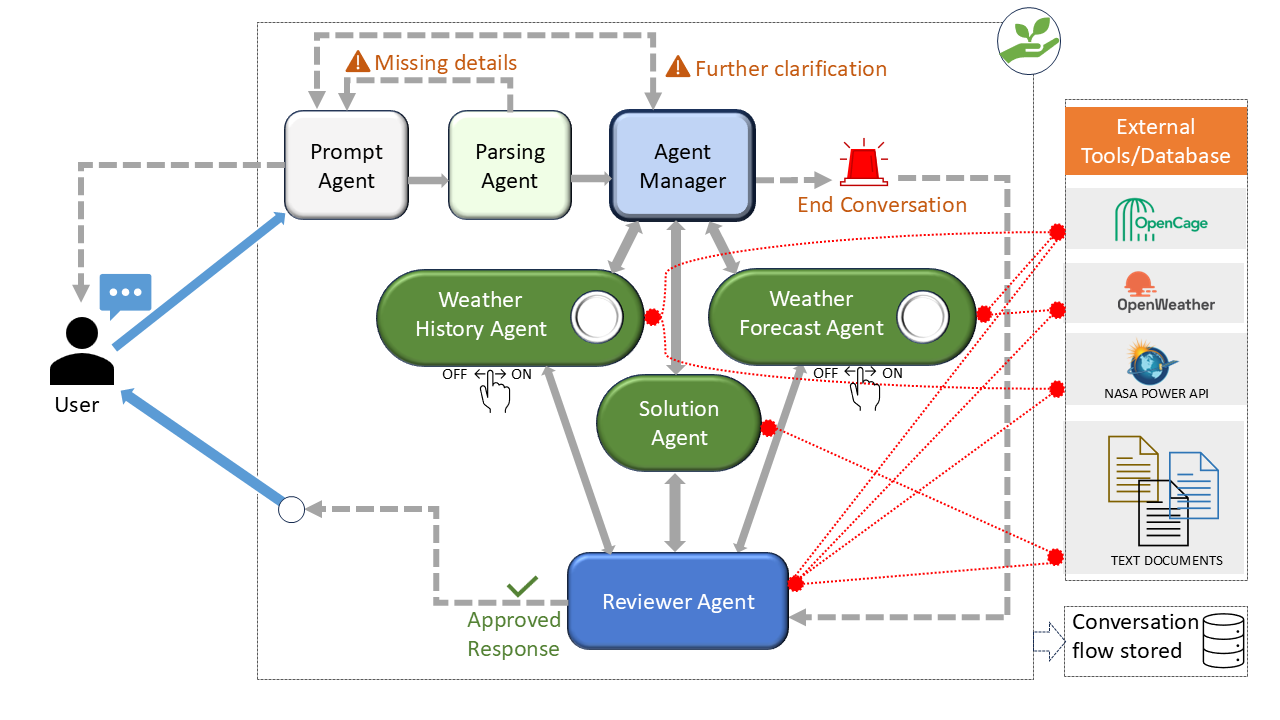}
    \caption{\textbf{AgroAskAI} Multi-Agent Architecture}
    \label{fig:architecture}
\end{figure*}

\section{Design and Methodology}

We developed a framework grounded in the agentic AI paradigm to address farmer queries related to agriculture, as illustrated in Figure~\ref{fig:architecture}. The framework comprises a collection of specialized agents, each designed to perform distinct tasks within a well-defined logical workflow. In this section, we provide a detailed description of each agent, its role within the overall framework, as well as the inter-agent connectivity and the structured exchange of information between agent pairs. The pseudocode for the agentic components is deferred to Appendix~A.

\paragraph{Prompt Agent.} Serves as the primary interface with the user, initiating conversations with a greeting and collecting user input. It forwards this input to the Parsing Agent for further processing, while also managing follow-up interactions. This includes handling requests for missing or ambiguous information from other agents, prompting the user for clarification, or notifying them when a query cannot be resolved. A key responsibility of the Prompt Agent is to identify the language used by the user, ensuring all subsequent interactions occur in that same language to maintain a seamless and personalized experience.

\paragraph{Parsing Agent.} Responsible for interpreting the user's input received from the Prompt Agent. It extracts essential information, including user intent, geographic location, and relevant time frame, to build a localized and actionable context. When key details are missing or the intent is unclear, the Parsing Agent collaborates with the Prompt Agent to request clarification, ensuring that the system maintains both accuracy and relevance in its responses. Upon complete information, the structured (JSON) data and the user query are then passed to the Agent Manager for further processing. 

\paragraph{Agent Manager.} Acts as the central coordinator within the system, interpreting the localized context and autonomously selecting the appropriate specialized agents to activate. By invoking only the relevant agents, it ensures efficient and focused task execution. It also plays a critical role in aggregating the information needed by the Solution Agent to craft a complete and contextually accurate response. Once a satisfactory response is finalized, typically following reviews from the Reviewer Agent, the Agent Manager terminates the response generation cycle and signals the Reviewer Agent to send the final solution.

\paragraph{Weather Forecast Agent.} Specializes in handling queries related to future or predicted weather conditions, utilizing external services such as OpenCage\footnote{OpenCage Geocoding API: https://opencagedata.com} and OpenWeather\footnote{OpenWeather Forecasting API: https://openweathermap.org} to retrieve location-specific forecast data. By integrating these services, the agent provides timely and relevant climate insights to support user decision-making.

\paragraph{Weather History Agent.} Designed to handle inquiries about past weather patterns, the Weather History Agent retrieves historical climate data using tools like OpenCage for geospatial resolution and the NASA POWER API\footnote{NASA POWER API: https://power.larc.nasa.gov} for reliable weather records. This allows the extraction of historical climate trends to support decision-making.

\paragraph{Solution Agent.} Generates context-aware, actionable responses for queries that require recommendations on agricultural practices, specifically in climate-sensitive agricultural settings. It draws upon information provided by the Agent Manager and consults external text documents to supplement domain-specific knowledge.

\paragraph{Reviewer Agent.} Plays a critical role in assessing responses generated by the Solution Agent across multiple dimensions, including clarity, technical soundness, fairness, and alignment with user intent. When necessary, it consults external sources to detect and reduce hallucinations in Solution Agent's responses. If the output is found to be incomplete or suboptimal, the Reviewer Agent can trigger a new cycle of agent selection and response generation via the Agent Manager. This iterative feedback mechanism ensures the final output is accurate, contextually appropriate, and socially responsible.

\subsection{Model Specification}
AgroAskAI is built using the Microsoft Semantic Kernel framework and utilizes OpenAI's GPT-4 Learning Language Model (LLM) to understand the natural language of user queries. The architecture of AgroAskAI is inspired by agentic AI design principles, such as perception stage, goal decomposition, and safety guardrails. This system consists of multiple agents whose tasks include interacting with user using the language of interest, parsing user input for necessary information, conducting agent selection, retrieving necessary external data, generating novel solutions and adaptations for user input. AgroAskAI also logs the complete conversation flow between agents for each user query, enabling traceability, transparency, and future system improvements.
\begin{figure*}[!h]
    \centering
    \includegraphics[scale=0.53]{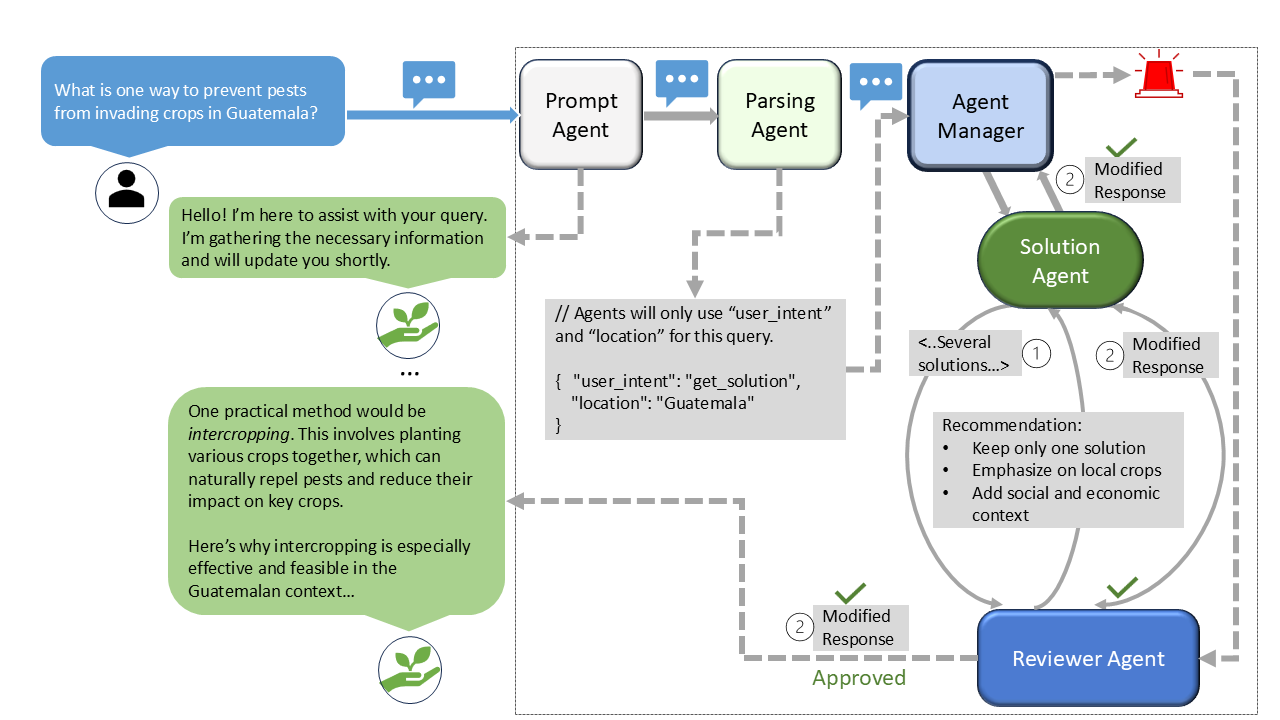}
    \caption{A sample decision-support by \textbf{ AgroAskAI.}}
    \label{fig:conversation}
\end{figure*}

Data is sourced through a Retrieval-Augmented Generation (RAG) strategy~\cite{singh2025agentic}, which allows the system to access up-to-date and context-specific data from external APIs. These sources include, but are not limited to, NASA’s POWER Project for historical climate data, OpenWeather for forecasts, and OpenCage for location-based geocoding. These components are orchestrated through a centralized control logic (Agent Manager) that activates the right agents for each task based on user input. This design enables the system to remain flexible, context-aware, and efficient in handling diverse climate-related queries.

Figure~\ref{fig:conversation} illustrates a sample conversation on ``What is one way to prevent pests from invading crops in Guatemala?''~\cite{hart2017guatemala}. Upon receiving the query, the prompt agent extracts key details (e.g., user\_intent, location) and forwards them, along with the original question, to the Agent Manager. Recognizing that no weather data is required, the Agent Manager directs the request to the solution agent. The solution agent generates an initial answer using relevant tools, after which the reviewer agent evaluates its strengths and shortcomings—particularly noting that the user requested a single solution and that local crops and socio-economic context should be specified. The solution agent incorporates these recommendations, and the revised answer is approved. If approval was withheld, the solution agent would continue iterating until a satisfactory solution was achieved. Upon approval, the Agent Manager ends the conversation, and the final response is delivered to the user. The complete conversation and response details are provided in Appendix~B.

\begin{table*}[!h]
\begin{tabular}{p{0.22\linewidth} | p{0.18\linewidth}|p{0.24\linewidth} | p{0.25\linewidth}}

\textbf{} & 
  \textbf{CROPWAT} & 
  \textbf{ChatGPT} & 
  \textbf{AgroAskAI} \\ \hline
\vspace{0.5\baselineskip}\textbf{Weather Information} & 
\begin{itemize}
    \item Detailed rainfall metrics
    \item No temperature metrics
\end{itemize} & 
  \begin{itemize}
      \item Listed average rainfall and historical weather conditions during 3 out of 20 trial runs and listed forecasts (temperature and precipitation) during 7 runs.
      \item Listed examples of climate issues during 7 out of the 20 runs.
  \end{itemize}
 & 
\begin{itemize}
      \item  Listed average rainfall and historical weather conditions during 16 of the trial runs and listed forecasts (temperature and precipitation) during 16 trial runs.
      \item  Listed examples of climate issues during 11 runs.
  \end{itemize} \\ \hline
\textbf{Water Conservation} & Outputs quantified irrigation scheduling and field efficiency metrics
 & 
 Provides implementation steps during 0 out of 20 runs & Mentions different harvesting system types during 15 runs. Provides implementation steps during 13 runs\\ \hline
\textbf{Livestock advice} & 
  Not addressed & 
  Addressed in 16 trial outputs & 
  Addressed in all 20 trial outputs \\ \hline
\textbf{Soil Conservation} & 
  Not addressed & 
  Names the methods of conservation in 12 outputs
 & 
 Names the methods of conservation in all 20 outputs. Out of which, 13 outputs had implementation instructions for conservation \\ \hline
\textbf{Adaptability to Farm Size/Type} & 
  Gives quantified data for optimal timing irrigation strategies but does not list them using practical language. & 
  9 outputs provided solutions accounting for farm-specific conditions (e.g., small vs. large-scale water management). & 
  16 outputs offered small and large-scale solutions with step-by-step instructions for varying farm conditions   (e.g., barrels vs tanks in water conservation) \\ \hline
\textbf{Inclusion of Support Programs or Resources} & 
  Not addressed & 
  4 outputs mentioned specific projects and institutional support (e.g. KCSAP, NDMA) & 
  16 outputs mentioned specific projects and institutional support\\ \hline  
\textbf{Multilingual Support} & Officially provides multilingual support for 4 languages & Officially provides multilingual read/write support for 50+ languages & Officially provides multilingual read/write support for 50+ languages.\\ \hline
\end{tabular}\caption{{Comparison of CROPWAT, ChatGPT, and AgroAskAI based on 20 different user queries}. The comparison highlights the enhanced capabilities of AgroAskAI in providing detailed responses.}\label{table1}
\end{table*}
\section{Experimental Results}
To evaluate AgroAskAI’s effectiveness, reliability, multi-agent, and multilingual capabilities in supporting climate adaptation decisions for agriculture, we curated region-specific farmer queries and asked AgroAskAI to provide decision-support. Our evaluation focused on queries representative of the challenges faced by rural farming communities, including drought planning, irrigation strategies, crop selection, and pest control. The key highlights are:
\begin{enumerate}
    \item \textbf{Comparison of AgroAskAI with Other Models} to evaluate AgroAskAI's response against a single-agent large language model (ChatGPT) and a traditional rule-based decision tool (CROPWAT).
    \item \textbf{Importance of Agent Manager}: in selecting and orchestrating the flow of reasoning among specialized agents.
    \item \textbf{Significance of the Reviewer Agent}: operating as an internal governance layer within the system.
    \item \textbf{Handling Missing Details}: Tackling under-specified queries and missing information.
    \item \textbf{Multi-lingual Capabilities} of AgroAskAI enabled by the underlying large language model.
\end{enumerate}

\paragraph{Comparison with Other Models} To assess the quality of AgroAskAI's responses to agriculture-related queries, the responses are compared against (1)~ChatGPT and (2)~CROPWAT (\url{https://www.fao.org/land-water/databases-and-software/cropwat/en/}). Interactions with ChatGPT were conducted using the OpenAI Chat GPT web interface during June-August 2025. The model used for all experiments was GPT-5.1 (the default model available on the ChatGPT platform at that time). No system prompt modifications, custom instructions, or fine-tuning were applied beyond ChatGPT’s default settings. Experiments using CROPWAT were performed with the latest available version as of July 2025, which was CROPWAT 8.0 (FAO). Agrifriendly could not be included in the comparison due to the lack of access to its source code or API. Also, Farmer.Chat required users to either upload proprietary content or join an existing registered team, restricting independent evaluation of the system. 

Table~\ref{table1} summarizes the comparison of three models evaluated across six dimensions: weather information, water conservation, livestock advice, soil conservation, farm adaptability, support resources, and multi-lingual capabilities, based on 20 trial inputs.

For example, when asked about water management strategies in Kitui, Kenya, ChatGPT briefly mentioned \textit{zai pits} and \textit{mulching} without elaboration, while CROPWAT focused solely on irrigation metrics. In contrast, AgroAskAI detailed specific water harvesting systems and provided clear instructions on how to apply them.


Regarding weather information, all three systems provided average rainfall data, but only AgroAskAI and ChatGPT included contextual climate risks like dry spells and heat. AgroAskAI additionally offered temperature data (21–27°C), missing in CROPWAT.


Regarding livestock advice, CROPWAT was unable to address any concerns, while AgroAskAI recommended drought-resilient breeds and integrated crop-livestock systems, which were often missing in ChatGPT's response. Similarly, for soil conservation, AgroAskAI explained practices like composting and cover cropping in practical terms, while ChatGPT listed techniques with minimal detail and CROPWAT lacked relevant content. 
Regarding adaptability to different farm sizes or types, AgroAskAI tailored advice to both small and large-scale operations. It offered concrete differences in implementation (e.g., tanks vs. barrels), whereas ChatGPT remained general, and CROPWAT provided timing metrics but no practical adaptation suggestions. In terms of institutional support, ChatGPT listed several programs like the Kenya Climate Smart Agricultural Project (KCSAP) and the National Drought Management Authority Act (NDMA), while AgroAskAI and CROPWAT did not reference any specific support systems. This suggests that integration with external programs remains a gap in the current system. Finally, both ChatGPT and AgroAskAI provide broader multi-lingual support than CROPWAT, which is restricted to four languages only. 

Overall, AgroAskAI consistently produced actionable, context-aware responses, outperforming both ChatGPT and CROPWAT across multiple dimensions. Its outputs were structured as step-by-step recommendations written in clear, practical language, enabling farmers and extension workers to directly apply the advice in field conditions. In contrast, ChatGPT often provided generic, surface-level lists with limited contextual agronomic grounding, while CROPWAT, though precise in its numerical calculations, offered quantified results without operational guidance. For smallholder farmers, the gap between information and actionable decision support is often where existing digital tools fall short. AgroAskAI bridges this gap by combining agronomic reasoning, localized context, and practical instructions, transforming analytical outputs into ready-to-use climate-adaptive recommendations. Appendix~C contains the outputs of CROPWAT, ChatGPT, and AgroAskAI for a particular query.

\paragraph{Importance of Agent Manager} To empirically evaluate the impact of the Agent Manager on task relevance and interaction coordination in a multi-agent system, we generated various queries that may or may not require invoking the weather forecast or the weather history agent. The Agent Manager successfully orchestrated the specialized agents to obtain an appropriate context-relevant response. For instance, in the sample query demonstrated in Figure~\ref{fig:conversation}, the Agent Manager activates only the Solution Agent since the other agents (weather history and weather forecast) are not required for the response. We provide a couple of additional examples. In Example 1, the Agent Manager invokes the Weather History Agent only, and in Example 2, only the Weather Forecast Agent is invoked. The Agent Manager successfully terminated the conversation once an appropriate solution has been generated.
\small{
\begin{tcolorbox}[title=Example 1. Agent Manager invokes Weather Forecast, colback=gray!5!white, colframe=black!75!black, sharp corners]
  \textbf{User~\cite{lopez2019food}:} What is the predicted rainfall in San Marcos, Guatemala, tomorrow? \\
  \textbf{AgroAskAI:} \textit{Hello, I'm here to assist with your query. I'm gathering the necessary information and will update you shortly.}\\
  \textbf{AgroAskAI:} \textit{The approved forecast indicates that in San Marcos, Guatemala, on August 12, 2025, there will be moderate rain with a predicted rainfall of $13.71$ mm. The daytime temperature will be approximately $25.19$°$C$, and at night, it will drop to around $19.18$°$C$. Additional details include $74\%$ humidity, wind speeds of $12.7$ kph from the southwest ($213$°$C$), and $64\%$ cloud cover. Sunrise is at 07:52, and sunset is at 20:32.}

\end{tcolorbox}

\begin{tcolorbox}[title=Example 2. Agent Manager invokes Weather History, colback=gray!5!white, colframe=black!75!black, sharp corners]
  \textbf{User~\cite{mutunga2017smallholder}:} How has the average temperature in Kitui, Kenya, changed over the past 20 years?\\
\textbf{AgroAskAI:} \textit{Hello, I'm here to assist with your query. I'm gathering the necessary information and will update you shortly.}\\
  \textbf{AgroAskAI:} \textit{From $2005$ to $2025$ in Kitui, Kenya, the annual average temperature showed a slight increase. The data indicates a gradual fluctuation within annual averages, starting at around $23.62$°$C$ in $2003$ and reaching $23.83$°$C$ in $2023$, with some years experiencing notable variations. Generally, this period reflects relatively stable but slightly warming trends in temperature.}
 \end{tcolorbox}
 }

\paragraph{Significance of the Reviewer Agent}  
To evaluate the Reviewer Agent, we compared AgroAskAI with a modified version without any Reviewer Agent. AgroAskAI consistently ensures factual accuracy and alignment with user intent in all the queries, which are often missed in the responses by a similar model without any Reviewer Agent. For example, while answering the query: ``What crops will survive the upcoming drought in Kitui, Kenya?'' \cite{mutunga2017smallholder}, the model without Reviewer Agent lists adaptations irrelevant to the user prompt, such as solutions to temperature changes and tips for water conservation. Such hallucinations are not found in the responses from AgroAskAI.

\paragraph{Handling Missing Details} We measure the effectiveness of AgroAskAI in detecting and resolving underspecified or incomplete queries. We found that the Parse Agent is always able to detect missing information related to location and timestamp when those information are relevant for the query. The Parse Agent also notifies the user when the intent is out of scope for AgroAskAI to answer. Moreover, Agent Manager sometimes requires more details and notifies the Prompt Agent to gather further clarification. Example 3 illustrates the effectiveness of AgroAskAI in identifying missing details.
\begin{tcolorbox}[title=Example 3. Missing location information, colback=gray!5!white, colframe=black!75!black, sharp corners]
  \textbf{User:} What are the predicted weather conditions and what can I do to manage my crops? \\
  \textbf{Prompt Agent to User:} \textit{Hello! I’m here to assist with your query. I’m gathering the necessary information and will update you shortly.}\\
  \textbf{Parse Agent:} 
  \begin{verbatim}
{   "user_itent": "get_solution",
    "location": "null",      
    "start_year": "", 
    "end_year": "", 
    "forecast_date": ""
}
\end{verbatim}\\
  \textbf{Prompt Agent to User}: Please enter a location (e.g., city, state, or country):\\
  \textbf{User:} Kitui, Kenya\\
  \textbf{Prompt Agent to Parse Agent:} Updated location to Kitui, Kenya.\\
  $<$response generation continues$>$
\end{tcolorbox}

\paragraph{Multi-lingual Capabilities} AgroAskAI supports queries in multiple languages, utilizing the capabilities of the large language models. AgroAskAI's Prompt Agent identifies the language in which the user asks a query and ensures that subsequent interactions with the user are in the same language. For example, for the query, \textit{``hali ya hewa ikoje leo katika kitui kenya''} (``What is the predicted weather in Kitui, Kenya?''), AgroAskAI correctly identifies the language as Swahili and responds:
\begin{tcolorbox}\begin{itemize}
  \item \textbf{Muhtasari:} Anga lina mawingu kidogo na vipindi vya uwazi.
  \item \textbf{Joto:} Mchana ni 28.57°C, na usiku ni 21.69°C.
  \item \textbf{Mvua:} Hakuna mvua inayotarajiwa.
  \item \textbf{Unyevu:} 28\%.
  \item \textbf{Upepo:} Kasi ya upepo ni 21.6 km/h, ukielekea 138°.
  \item \textbf{Kiwango cha UV:} 11.62 (Juu sana).
  \item \textbf{Jua:} Linachomoza saa 5:30 asubuhi na kuzama saa 5:34 jioni.
\end{itemize}
\end{tcolorbox}

Overall, AgroAskAI demonstrated superior performance by delivering localized, adaptable, and step-by-step guidance, bridging the gap between quantitative tools like CROPWAT-8.0 and generic generative models like ChatGPT-5.1.

\section{Conclusion and Discussion}
In the face of providing climate-adaptive decision support to farmers---ranging from prolonged droughts to unpredictable rainfall---AgroAskAI provides detailed step-by step guidance combining precision of domain-specific tools with the power of generative AI, and leveraging a multi-agent, role-specialized architecture with built-in governance mechanisms. Experimental results affirm AgroAskAI’s superior performance in delivering actionable and localized recommendations compared to both static modeling tools and generic single-agent AI models. As agriculture faces an increasingly uncertain climate future, AgroAskAI exemplifies how agentic AI can drive sustainable, scalable, and accountable adaptation strategies, empowering farmers to thrive in a changing world.

\paragraph{Limitations.} While AgroAskAI shows promising results in our empirical evaluation, we note that the system's performance is bounded by the quality and scope of external data sources it integrates, including weather databases and local document repositories. Moreover, its multilingual capabilities, while strong, rely on the underlying language model (OpenAI) and may require further fine-tuning for under-resourced dialects. Finally, real-world deployment in low-connectivity environments remains an operational challenge, requiring lightweight interfaces and offline capabilities. One \textit{potential solution is the establishment of centralized computing hubs, such as community centers, where a technically proficient individual can operate AgroAskAI on behalf of multiple farmers}. This shared-access model can bridge digital gaps while ensuring that even farmers without personal devices or internet connectivity benefit from the system’s capabilities.

\paragraph{Future Work.} Going ahead, we will focus on extensive user testing with smallholder farmers to evaluate the usability, accessibility, and cultural appropriateness of AgroAskAI. This includes participatory design sessions, where farmers interact and critique system outputs; field trials to assess the clarity, trustworthiness, and impact of recommendations; and iterative refinement based on real-world feedback. Special attention will be given to testing across regions with varying literacy levels, dialects, and technological familiarity, ensuring that the interface, language, and decision-support outputs are truly aligned with farmers' needs.

\section*{Acknowledgements} N.C. gratefully acknowledges the support of Douglass WiSE’s SUPER Research Program.
\bibliography{references}

@article{hoa2022exploring,
  author    = {Hoa Sen, L. T. and Bond, J. and Hoang, H. D. T.},
  title     = {Exploring smallholder farmers’ climate adaptation decision-making in mountainous areas of Central Vietnam: implications for extension services},
  journal   = {The Journal of Agricultural Education and Extension},
  volume    = {29},
  number    = {2},
  pages     = {247--268},
  year      = {2022},
  doi       = {10.1080/1389224X.2022.2039248},
  url       = {https://doi.org/10.1080/1389224X.2022.2039248}
}

@article{lamichhane2022nepal,
title = {What motivates smallholder farmers to adapt to climate change? Insights from smallholder cropping in far-western Nepal},
journal = {Anthropocene},
volume = {40},
pages = {100355},
year = {2022},
issn = {2213-3054},
doi = {https://doi.org/10.1016/j.ancene.2022.100355},
author = {Prahlad Lamichhane and Kelly K. Miller and Michalis Hadjikakou and Brett A. Bryan}
}

@article{zagre2024senegal,
  
AUTHOR={Zagre, Inoussa  and Akinseye, Folorunso Mathew  and Worou, Omonlola Nadine  and Kone, Mama  and Faye, Aliou },
         
TITLE={Climate change adaptation strategies among smallholder farmers in Senegal’s semi-arid zone: role of socio-economic factors and institutional supports},
        
JOURNAL={Frontiers in Climate},
        
VOLUME={Volume 6 - 2024},

YEAR={2024},

URL={https://www.frontiersin.org/journals/climate/articles/10.3389/fclim.2024.1332196},

DOI={silici2021adaptive}
}

@article{silici2021adaptive,
  author    = {Silici, Laura and Rowe, Andy and Suppiramaniam, Nanthikesan and Knox, Jerry W.},
  title     = {Building adaptive capacity of smallholder agriculture to climate change: Evidence synthesis on learning outcomes},
  journal   = {Environmental Research Communications},
  volume    = {3},
  number    = {12},
  pages     = {122001},
  year      = {2021}
}

@article{raza2025trism,
  title={Trism for agentic ai: A review of trust, risk, and security management in llm-based agentic multi-agent systems},
  author={Raza, Shaina and Sapkota, Ranjan and Karkee, Manoj and Emmanouilidis, Christos},
  journal={arXiv preprint arXiv:2506.04133},
  year={2025}
}

@article{castellanos2019climate,
  title={Climate change, agriculture, and adaptation options for Guatemala},
  author={Castellanos, Edwin J and Thomas, Timothy S and Dunston, Shahnila},
  year={2019},
  publisher={International Food Policy Research Institute}
}

@article{waddick2017effects,
  title={Effects of Climate Change on Agriculture in Guatemala and Honduras},
  author={Waddick, Karissa},
  journal={Global Majority E-Journal},
  volume={8},
  number={2},
  pages={109--120},
  year={2017}
}

@article{mutunga2017smallholder,
  title={Smallholder farmers perceptions and adaptations to climate change and variability in Kitui county, Kenya},
  author={Mutunga, Evelyn J and Charles, KN and Patricia, M},
  year={2017},
  publisher={OMICS International}
}

@article{singh2025agentic,
  title={Agentic retrieval-augmented generation: A survey on agentic rag},
  author={Singh, Aditi and Ehtesham, Abul and Kumar, Saket and Khoei, Tala Talaei},
  journal={arXiv preprint arXiv:2501.09136},
  year={2025}
}

@misc{charlton2023climate,
  author       = {Charlton, Emma},
  title        = {This is what the climate crisis is costing economies around the world},
  year         = {2023},
  month        = {November},
  url          = {https://www.weforum.org/stories/2023/11/climate-crisis-cost-global-economies/},
  note         = {Accessed: 2025-06-10}
}

@misc{smith2023historic,
  author       = {Smith, Adam B.},
  title        = {2023: A historic year of U.S. billion-dollar weather and climate disasters},
  year         = {2024},
  month        = {January},
  url          = {https://www.climate.gov/news-features/blogs/beyond-data/2023-historic-year-us-billion-dollar-weather-and-climate-disasters},
  note         = {Accessed: 2025-06-10}
}

@misc{hart2017guatemala,
  author       = {Hart, Nancy},
  title        = {At Mexico-Guatemala Border, Sterile Insect Technique has Controlled the Northward March of Medflies Since 1982},
  year         = {2017},
  month        = {November},
  url          = {https://www.iaea.org/newscenter/news/at-mexico-guatemala-border-sterile-insect-technique-has-controlled-the-northward-march-of-medflies-since-1982.},
  note         = {Accessed: 2025-06-10}
}

@misc{farmbeats,
  author       = {Microsoft},
  title        = {FarmBeats: AI, Edge \& IoT for Agriculture},
  year         = {2019},
  url          = {https://www.microsoft.com/en-us/research/project/farmbeats-iot-agriculture/},
  note         = {Accessed: 2025-06-08}
}

@article{wangfuture,
  title={Future-proof finance: Navigating climate risks and ESG goals with agentic AI and collaborative foresight},
  author={Wang, Francis and Chen, Tao and Koh, Lionel and Wu, Amanda},
  year={2025},
  journal={FW. VISION Futures Thinking Inc.}
}

@article{gujjar2025agri,
  title={Agri Friendly Conversational AI Chatbot Using Open Source Framework},
  author={Gujjar, JP and Kumar, HRP},
  journal={Indian Journal of Science and Technology},
  volume={18},
  number={8},
  pages={580--585},
  year={2025}
}

@article{mohammedagentic,
  title={Agentic Ai Forging Adaptive, Equity-Driven Governance Pathways for Sustainable Futures},
  author={Mohammed Salah, A and Alnoor, Alhamzah and Abdelfattah, Fadi and Chew, Dr XinYing},
  journal={XinYing, Agentic Ai Forging Adaptive, Equity-Driven Governance Pathways for Sustainable Futures},
year={2025}
}

@article{singh2024farmer,
  title={Farmer. Chat: Scaling AI-Powered Agricultural Services for Smallholder Farmers},
  author={Singh, Namita and Wang'ombe, Jacqueline and Okanga, Nereah and Zelenska, Tetyana and Repishti, Jona and Mishra, Sanjeev and Manokaran, Rajsekar and Singh, Vineet and Rafiq, Mohammed Irfan and Gandhi, Rikin and others},
  journal={arXiv preprint arXiv:2409.08916},
  year={2024}
}

@article{tonmoyagribuddy,
  title={AgriBuddy: An Agentic AI System for Bangladeshi Agriculture Using RAG and Vision Models},
  author={Tonmoy, Md Shaleh Islam and Rabbi, Rawhatur and Hassan, Md Rezuwan and Naswan, Ruwad and Shome, Tanmoy},
year = {2025}
}

@article{lopez2019food,
  title={Food security and agriculture in the Western Highlands of Guatemala},
  author={Lopez-Ridaura, Santiago and Barba-Escoto, Luis and Reyna, Cristian and Hellin, Jon and Gerard, Bruno and Van Wijk, Mark},
  journal={Food Security},
  volume={11},
  number={4},
  pages={817--833},
  year={2019},
  publisher={Springer}
}
\clearpage
\appendix
\section{Pseudo Code for Agents in AgroAskAI}\label{sec:pseudocode}


\begin{algorithm}[!h]
\caption{PromptAgent.invoke}
\label{alg:prompt-agent}
\begin{algorithmic}[1]
\REQUIRE User query $\texttt{q}$ and clarification request $\texttt{r}$
\STATE $\texttt{kb} \gets \textit{OpenAIChatCompletion}(\texttt{GPT-4o})$
\STATE \textbf{Initiate} \texttt{language} = ``English''
\IF{$\texttt{q} \neq \texttt{null}$}
    \STATE \texttt{language} $\gets$ \texttt{kb}.{extract\_language}(\texttt{q})
    \STATE \textbf{Display} $\texttt{greeting\_message}(\texttt{language})$
    \STATE ParseAgent.invoke(\texttt{q})
\ELSE
    \STATE \texttt{qr} $\gets$ \texttt{kb}.{generate\_clarification}(\texttt{r})
    \STATE \textbf{Display} $\texttt{qr}$ to user
    \STATE \textbf{Wait for} $\texttt{user\_input}$
    \STATE ParseAgent.invoke(\texttt{user\_input})
\ENDIF
\end{algorithmic}
\end{algorithm}

\begin{algorithm}[H]
\caption{ParseAgent.invoke}
\label{alg:parse-agent}
\begin{algorithmic}[1]
\REQUIRE \texttt{nl\_q}: Processed natural language query from the Prompt Agent
\STATE \texttt{kb} $\gets$ \textit{OpenAIChatCompletion}(\texttt{GPT-4o})
\STATE \texttt{i} $\gets$ \texttt{kb}.extract\_user\_intent(\texttt{nl\_q})
\IF{!\texttt{kb}.is\_in\_scope($i$)}
\STATE \texttt{cr} $\gets$ \texttt{kb}.\textit{notify\_out\_of\_scope}(\texttt{i})
\STATE PromptAgent.invoke(null,\texttt{cr})
\RETURN
\ENDIF
\STATE \texttt{mv} $\leftarrow$ null $\quad\quad\quad\quad\quad\quad\quad\quad\quad\quad$ \COMMENT{missing values} 
\IF{\texttt{kb}.is\_location\_req(\texttt{i})}
\STATE \texttt{l} $\gets$ \texttt{kb}.extract\_location(\texttt{nl\_output})
\IF{\texttt{l}==null}
\STATE \texttt{mv} $\leftarrow$ \texttt{mv} $\cup$ \texttt{\{"l"\}}
\ENDIF
\ENDIF
\IF{\texttt{kb}.is\_timerange\_required(\texttt{i})}
\STATE \texttt{t = [t1,t2]}$\gets$\texttt{kb}.extract\_time\_range(\texttt{nl\_output})
\IF{\texttt{t}==null}
\STATE \texttt{mv}$\leftarrow$\texttt{mv} $\cup$ \texttt{\{"t"\}}
\ENDIF
\ENDIF

\STATE \texttt{parsed} $\gets$to\_JSON(\texttt{i}, \texttt{l}, \texttt{t})
\WHILE{\texttt{mv}!=null}
    \STATE \texttt{cr} $\gets$ \texttt{kb}.\textit{generate\_clarification\_request}(\texttt{mv})
    \STATE \texttt{user\_input} $\gets$ PromptAgent.{invoke}(null, \texttt{cr})
    \STATE \texttt{[i,l,t]} $\leftarrow$ \texttt{kb}.Extract(\texttt{user\_input}, \texttt{parsed})
    \STATE \texttt{parsed} $\gets$to\_JSON(\texttt{i}, \texttt{l}, \texttt{t})
    \STATE \texttt{mv} $\leftarrow \{k\in\{\texttt{l, t}\}: k==null\}\}$ 
\ENDWHILE
\STATE AgentManager.invoke(\texttt{nl\_q, parsed})
\end{algorithmic}
\end{algorithm}

\begin{algorithm}[H]
\caption{AgentManager.invoke}
\label{alg:chat-manager}
\begin{algorithmic}[1]
\REQUIRE Query \texttt{nl\_q} and Parsed query \texttt{parsed}
\STATE \texttt{kb} $\gets$ \textit{OpenAIChatCompletion}(\texttt{GPT-4o})
\STATE \textbf{Specify} {termination\_strategy}
\STATE {termination()} = \texttt{kb}.{generate}(termination\_strategy)
\STATE \textbf{Specify} {selection\_strategy}
\STATE {selection()} = \texttt{kb}.{generate}(selection\_strategy)
\STATE \textbf{Initialize} \texttt{chat\_history} $\gets$ \texttt{nl\_q}
\STATE \textbf{Set} \texttt{agents = \{Prompt, Parse, Solution, WeatherHistory, WeatherForecast, Reviewer\}}
\STATE \textbf{Set} \texttt{current\_agent} $\leftarrow$ null
\STATE \textbf{Initiate} \texttt{conversation\_active} $\leftarrow$ True
\WHILE{\texttt{true}}
 \STATE \texttt{next\_agent} $\leftarrow $ selection()
    \IF{\texttt{next\_agent} == None}
        \STATE ReviewerAgent.invoke(null)
        \STATE \textbf{break}
    \ELSE
        \STATE \texttt{current\_agent} $\leftarrow$ \texttt{next\_agent}
    \ENDIF
    \STATE \texttt{out}$\leftarrow \texttt{current\_agent}.invoke$(\texttt{chat\_history})
    \STATE \texttt{chat\_history}.$append$(\texttt{out})
    \IF{termination\texttt{(chat\_history, out)}}
        \STATE \texttt{conversation\_active} $\leftarrow$ False
    \ENDIF
   
\ENDWHILE
\end{algorithmic}
\end{algorithm}

\begin{algorithm}[H]
\caption{WeatherForecastAgent.invoke}
\label{alg:forecast-agent}
\begin{algorithmic}[1]
\REQUIRE \texttt{parsed\_query}, $\mathit{get\_forecast\_data}$
\ENSURE \texttt{forecast\_summary}
\STATE \texttt{l} $\gets$ \texttt{parsed\_query}.\texttt{location}
\STATE \texttt{t} $\gets$ \texttt{parsed\_query}.\texttt{time}
\STATE \texttt{forecast\_data} $\gets$ $\mathit{get\_forecast\_data}(\texttt{l}, \texttt{t})$
\STATE \texttt{kb} $\gets$ $\mathit{OpenAIChatCompletion}(\texttt{GPT-4o})$
\STATE \texttt{res} $\gets$ \texttt{kb}.{GenerateSummary}(\texttt{forecast\_data})
\RETURN \texttt{res}
\end{algorithmic}
\end{algorithm}

\begin{algorithm}[H]
\caption{WeatherHistoryAgent.invoke}
\label{alg:history-agent}
\begin{algorithmic}[1]
\REQUIRE \texttt{parsed\_query}, $\mathit{get\_NASA\_data}$
\ENSURE \texttt{weather\_summary}
\STATE $\texttt{l} \gets \texttt{parsed\_query}.\texttt{location}$
\STATE $\texttt{t} \gets \texttt{parsed\_query}.\texttt{time}$
\STATE $\texttt{NASA\_data} \gets \mathit{get\_NASA\_data}(\texttt{l,t}
)$
\STATE $\texttt{kb} \gets \mathit{OpenAIChatCompletion}(\texttt{GPT-4o})$
\STATE \texttt{res} $\gets$ \texttt{kb}.{GenerateSummary}(\texttt{NASA\_data})
\RETURN \texttt{res}
\end{algorithmic}
\end{algorithm}

\begin{algorithm}[H]
\caption{SolutionAgent.invoke}
\label{alg:solution-agent}
\begin{algorithmic}[1]
\REQUIRE parsed\_query \texttt{p}, historical weather data \texttt{w}, forecast data \texttt{f}, $\mathit{get\_adaptations}()$
\ENSURE \texttt{solution\_plan}
\STATE \texttt{kb} $\gets \mathit{OpenAIChatCompletion}(\texttt{GPT-4o LLM})$
\STATE \texttt{data} $\gets \mathit{get\_adaptations}(p$.location)
\STATE \texttt{solution\_plan} $\gets$ null 
\STATE \texttt{approved} $\gets$ \text{False}
\WHILE{$\texttt{approved} == $False}
    \STATE \texttt{solution\_plan} $\gets$ \texttt{kb}.{GeneratePlan}(\texttt{data, p, w, f})
    \STATE \textbf{Send} \texttt{solution\_plan} to ReviewerAgent
    \STATE \texttt{res}$\gets$ ReviewerAgent.response(\texttt{solution\_plan})
    \STATE \textbf{Wait} for \texttt{approved} signal set by AgentManager
\ENDWHILE
\end{algorithmic}
\end{algorithm}

\begin{algorithm}[H]
\caption{ReviewerAgent.invoke}
\label{alg:reviewer-agent}
\begin{algorithmic}[1]
\REQUIRE parsed\_query \texttt{p}, previous agent output \texttt{a}
\ENSURE \texttt{reviewer\_output}
\STATE \texttt{kb} $\gets$ $\mathit{OpenAIChatCompletion}(\texttt{GPT-4o LLM})$
\STATE \texttt{f} $\gets$ $\mathit{get\_forecast\_data}(\texttt{p.location}, \texttt{p.time})$
\STATE \texttt{n} $\gets$ $\mathit{get\_NASA\_data}(\texttt{p.location}, \texttt{p.time})$
\STATE \texttt{q} $\gets$ $\mathit{get\_adaptations}(\texttt{p.location})$
\STATE \texttt{review\_result} $\leftarrow$ \texttt{kb}.{EvaluateAgentOutput}$(\texttt{a}, \texttt{f}, \texttt{n}, \texttt{q}, \texttt{p})$
\IF{$\texttt{review\_result}.\mathit{NotApproved}$}
    \STATE \texttt{reviewer\_output} $\gets$ \texttt{kb}.{ConstructImprovement}$(\texttt{a}, \texttt{review\_result})$
\ELSE
    \STATE \texttt{reviewer\_output} $\gets$ "This solution is completely approved."
\ENDIF
\STATE \textbf{Send} \texttt{reviewer\_output} to AgentManager
\end{algorithmic}
\end{algorithm}

\section{Sample Decision-Support by AgroAskAI}\label{sec:example}
We present the complete conversation flow between agents corresponding to Figure~\ref{fig:conversation} below.\\

\textbf{User Input~\cite{hart2017guatemala}:} \textit{“What is one way to prevent pests from invading crops in Guatemala?”}

\vspace{1em}

\textbf{Prompt Agent:} \textit{“Hello! I’m here to assist with your query. I’m gathering the necessary information and will update you shortly.”}

\vspace{1em}

\textbf{Parse Agent (Raw Response):}
\begin{verbatim}
{
    "user_intent": "get_solution",
    "location": "Guatemala",      
    "start_year": "", 
    "end_year": "", 
    "forecast_date": ""
} 
\end{verbatim}
\textit{// Agents will only use “user\_intent” and “location” for this query.}

\vspace{1em}

\textbf{Solution Agent (Response \textcircled{\raisebox{-0.9pt}{1}}{}):} \\
“One effective way to prevent pests from invading crops in Guatemala is \textbf{integrated pest management (IPM)}. This sustainable approach combines various agricultural practices to protect crops while minimizing environmental impact. Considering Guatemala's tropical climate and crop practices, here are some practical pest control methods:

\begin{itemize}
    
    \item \textbf{Crop Diversification:} Plant a variety of crops in the same field to break pest cycles and reduce the risk of widespread infestation.
    \item \textbf{Timely Planting and Harvesting:} Planting just before the beginning of rains or at optimal times for pest avoidance can reduce pest densities and their impact.
    \item \textbf{Rotate Crops:} Change crop types in different seasons. This practice is especially effective in disrupting pests that target specific crops.
    \item \textbf{Use of Biological Controls:} Introduce pest predators (e.g., lady beetles) or use bio-pesticides derived from natural resources, which are safer and environment-friendly.
    \item \textbf{Intercropping:} Planting companion crops that repel pests. For instance, marigolds are known to deter nematodes and other harmful pests.
    \item \textbf{Proper Sanitation:} Regularly remove crop residues and weeds to avoid providing breeding grounds for pests.
    \item \textbf{Planting Pest-resistant or Hybrid Varieties:} Use hybrid crop varieties that are better adapted to resist infestation.
    \item \textbf{Soil Conservation and Fertilizer Application:} Improve soil health with manure or compost, which supports healthier plants that are less vulnerable to pest invasion.
\end{itemize}

These techniques not only reduce pest risks but also promote sustainable agricultural practices suitable for Guatemala's ecological conditions. Combining these strategies can provide long-term protection for crops.”

\vspace{1em}

\textbf{Reviewer Agent (Review of Response \textcircled{\raisebox{-0.9pt}{1}}{}):} \\
{The SolutionAgent proposes an integrated pest management (IPM) strategy and lists several methods (e.g., crop diversification, timely planting, biological controls, intercropping, sanitation, pest-resistant crops, and soil conservation).}

\textbf{Strengths of the Response:}
\begin{itemize}
    \item \textbf{Addresses the User’s Context:} The response takes Guatemala’s tropical climate into account, offering regionally relevant and adaptive measures.
    \item \textbf{Presents Practical and Usable Methods:} Diverse techniques like intercropping, use of biological controls, and crop rotation are helpful and widely applicable.
    \item \textbf{Scientifically Valid:} The proposed strategies align with modern sustainable farming practices and IPM principles.
\end{itemize}

\textbf{Shortcomings \& Missing Aspects:}
\begin{itemize}
    \item \textbf{Fails to Focus on ``One'' Method:} The user explicitly asked for "one way" to prevent pest invasions, but the response offers multiple techniques without clearly identifying or explaining the most effective single solution.
    \item \textbf{Lacks Emphasis on Local Crops:} The response doesn’t highlight specific Guatemalan crops that might benefit from these pest prevention strategies.
    \item \textbf{No Mention of Social or Economic Context:} Solutions should factor in the economic realities of Guatemalan farmers, such as cost-effectiveness and resource availability for small-scale farming.
\end{itemize}

\textbf{Recommendations for Improvement:}
\begin{itemize}
    \item \textbf{Prioritize a Single Key Solution:} Focus on one standout approach (e.g., intercropping or crop rotation) and provide a detailed explanation of its advantages in Guatemala's context.
    \item \textbf{Incorporate Specific Examples or Local Relevance:} Mention specific pests plaguing crops in Guatemala and how the selected method mitigates those pests most effectively.
    \item \textbf{Address Feasibility Concerns:} Include practical aspects like cost, space, and labor efficiency to ensure recommendations are actionable for all farmers.
\end{itemize}

{Unless updated, this response does not fully satisfy the user's request due to its broad focus and lack of prioritization.}

\vspace{1em}

\textbf{Solution Agent (Revised Response \textcircled{\raisebox{-0.9pt}{2}}{}):} \\
“If we were to focus on one key and practical method to prevent pests from invading crops in Guatemala, an optimal choice would be \textbf{intercropping}. This sustainable agricultural practice involves planting various crops together, which can naturally repel pests and reduce their impact on key crops. Here’s why intercropping is especially effective and feasible in the Guatemalan context:\\

\textbf{Why Intercropping is Recommended:}
\begin{enumerate}
    \item \textbf{Natural Pest Repellence:} Certain companion crops, such as marigolds or garlic, release natural chemicals that deter common agricultural pests. Aromatic herbs like basil or mint can also be intercropped for pest control benefits.
    \item \textbf{Disruption of Pest Life Cycles:} By mixing crops in the same area, pests that target specific crop species are less likely to damage extensively. This reduces the risk of significant yield losses.
    \item \textbf{Better Resource Use:} For smallholder farmers, intercropping utilizes limited farm space more efficiently and diversifies outputs, which can help mitigate financial risks.      
    \item \textbf{Environmental and Economic Sustainability:} Reduces the need for chemical pesticides, which not only lowers costs but also minimizes environmental impact and promotes healthier soil.
    \item \textbf{Locally Relevant Crops:} For Guatemalan farmers, combining staple crops like maize with beans or chili peppers has both pest management and economic benefits. Beans, for example, can naturally fix nitrogen into the soil while maize provides structural support.
\end{enumerate}

\textbf{Practical Example:} \\
Farmers in Guatemala could plant maize (a staple crop) alongside squash and beans. This traditional “Three Sisters” intercropping method:
\begin{itemize}
    \item Naturally repels pests,
    \item Improves soil fertility, and
    \item Maximizes yield diversity.
\end{itemize}

\textbf{Additional Recommendations:}
\begin{itemize}
    \item \textbf{Regular Crop Rotation:} Rotate crops with non-host plants annually to further minimize pests’ adaptability.
    \item \textbf{Training and Awareness:} Farmers could benefit from local workshops or training sessions on effective intercropping techniques and suitable companion crops for their region's pest problems.
\end{itemize}

\textbf{In conclusion}, intercropping stands out as a simple, cost-effective, and sustainable method for pest management in Guatemala. This approach will not only reduce pest invasions but also align with local farming traditions and resource constraints.”

\vspace{1em}

\textbf{Reviewer Agent (Review of Response \textcircled{\raisebox{-0.9pt}{2}}{}):} \\
{This revisited response provides a precise solution that fully focuses on intercropping as the primary method, as per the user’s request for "one way" to prevent pests. It explains the advantages, contextual relevance to Guatemala, and specific applications such as the local “Three Sisters” cropping strategy. Additionally, it considers implementation feasibility for Guatemalan farmers.}\\

\textbf{“This solution is completely approved.”}

\section{Comparison of CROPWAT, ChatGPT, and AgoAskAI}\label{sec:comparison}
\begin{figure}[!h]
    \centering
    \includegraphics[scale=0.5]{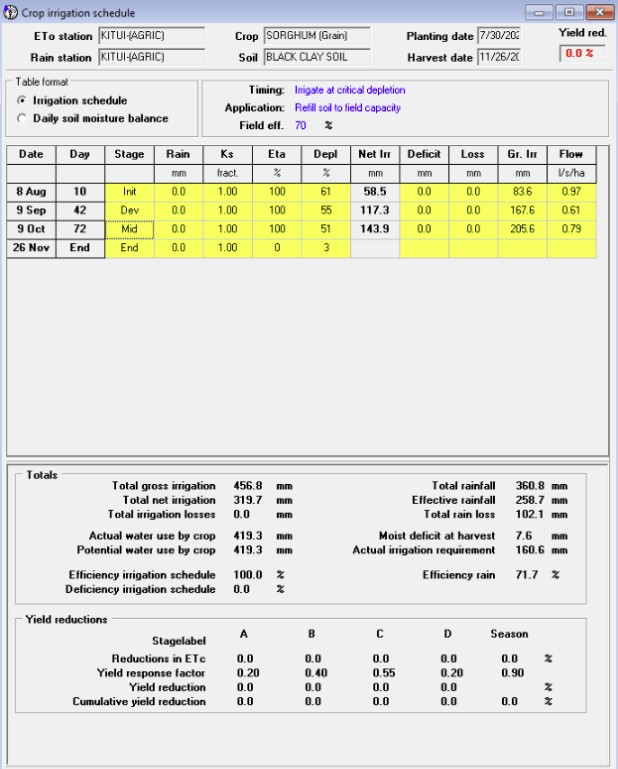}
    \caption{\textbf{CROPWAT's response.}}
    \label{fig:CROPWAT}
\end{figure}

We provide the response from AgroAskAI, ChatGPT, and CROPWAT when asked ``What are the climate issues in Kitui, Kenya and what can I do to protect my crops?''~\cite{mutunga2017smallholder}. Figure~\ref{fig:CROPWAT} shows the response of CROPWAT, where the rainfall metrics are clearly stated and information regarding ittigating scheduling is provided. However, the solution does not include temperature metrics nor livestock and soil conservation advice, and lacks specification of supporting programs.

\noindent The following is the response by ChatGPT.\\

\noindent \textbf{Climate Issues in Kitui, Kenya}
\begin{enumerate}
    \item \textbf{Unpredictable Rainfall Patterns}
\begin{itemize}
    \item Rainfall is erratic and unevenly distributed.
    \item The ``long rains'' (March--May) and ``short rains'' (October--December) are no longer reliable.
    \item Leads to poor planting seasons and crop failure.
\end{itemize}

\item \textbf{Frequent Droughts}
\begin{itemize}
    \item Kitui experiences recurrent droughts due to low annual rainfall (\textless{}700 mm in many areas).
    \item Droughts stress both crops and livestock, reducing yields and incomes.
\end{itemize}

\item \textbf{High Temperatures}
\begin{itemize}
    \item Rising average temperatures increase evapotranspiration.
    \item Crops require more water while water sources are shrinking.
\end{itemize}

\item \textbf{Land Degradation}
\begin{itemize}
    \item Deforestation and overgrazing have led to soil erosion.
    \item Soils are increasingly infertile and less capable of retaining moisture.
\end{itemize}

\item \textbf{Pest and Disease Outbreaks}
\begin{itemize}
    \item Warmer, wetter periods following droughts often bring outbreaks (e.g., fall armyworm).
    \item Climate change expands the habitat of crop pests.
\end{itemize}
\end{enumerate}
\vspace{1em}
\textbf{Adaptation Strategies for Farmers in Kitui}

\begin{enumerate}
\item \textbf{Adopt Climate-Resilient Crops}
\begin{itemize}
    \item Grow drought-tolerant and early-maturing varieties like:
    \begin{itemize}
        \item Sorghum
        \item Millet
        \item Pigeon peas
        \item Improved maize varieties (e.g., KDV4 or DK8031)
    \end{itemize}
\end{itemize}

\item \textbf{Use Water Conservation Techniques}
\begin{itemize}
    \item Zai pits or planting basins to trap water at the root zone.
    \item Mulching to reduce evaporation.
    \item Roof water harvesting and storage tanks for irrigation.
\end{itemize}

\item \textbf{Agroforestry}
\begin{itemize}
    \item Plant trees like Moringa, Neem, and \textit{Melia volkensii} to:
    \begin{itemize}
        \item Improve soil fertility
        \item Provide shade
        \item Increase biodiversity
        \item Act as windbreaks
    \end{itemize}
\end{itemize}

\item \textbf{Soil Conservation}
\begin{itemize}
    \item Use terracing, cover cropping, and composting to restore soil fertility.
    \item Encourage minimum tillage to preserve soil structure.
\end{itemize}

\item \textbf{Improved Farming Practices}
\begin{itemize}
    \item Use intercropping and crop rotation to reduce pest pressure.
    \item Follow seasonal climate forecasts to decide when and what to plant.
    \item Join farmer cooperatives or digital platforms for training and updates.
\end{itemize}

\item \textbf{Insurance \& Risk Management}
\begin{itemize}
    \item Enroll in index-based crop insurance (e.g., via ACRE Africa).
    \item Use mobile platforms (like iShamba) for weather alerts and advice.
\end{itemize}
\end{enumerate}
\noindent \textbf{Examples of Projects in Kitui Supporting Farmers}
\begin{itemize}
    \item \textbf{Kenya Climate Smart Agriculture Project (KCSAP)} – trains farmers on sustainable practices.
    \item \textbf{NDMA (National Drought Management Authority)} – offers early warning systems and response plans.
    \item \textbf{FAO \& World Bank programs} – promote soil restoration and water harvesting.\\
\end{itemize}

\noindent The following is the response by AgroAskAI.\\

\noindent {Based on the data available for Kitui, Kenya, here is a detailed and practical agricultural adaptation strategy tailored to the region:}

\vspace{1em}
\textbf{Observations from Historical Weather Data:}
\begin{enumerate}
    \item \textbf{Rainfall}: Kitui experiences low and irregular rainfall, ranging approximately from 0.08 mm to 7.33 mm monthly during the years analyzed (2010--2022). The rains are bimodal, with most of the heavier precipitation occurring in April, May, October, and November.
    \item \textbf{Temperatures}: Average temperatures remain relatively high, ranging from 21°C to 27°C with significant heat stress during certain months.
\end{enumerate}

\vspace{1em}
\textbf{Recommendations for Farmers in Kitui:}

To cope with Kitui's dry climatic conditions and make the most of sporadic rains:
\begin{enumerate}
\item \textbf{Adopt Drought-Resistant Crops}
\begin{itemize}
    \item Grow crops like sorghum, millet, and cowpeas, which are well-adapted to Kitui's arid climate.
    \item Utilize hybrid and drought-resistant seed varieties to increase tolerance to dry conditions.
    \item Diversify crop choices to spread risks and improve food security.
\end{itemize}

\item \textbf{Plant Timing and Early Maturing Crops}
\begin{itemize}
    \item \textbf{Plant just before the rains}: Monitor weather forecasts and prepare to sow seeds shortly before expected rainfall.
    \item Choose early-maturing crops to complete growth cycles during rainy periods and reduce water demand.
\end{itemize}

\item \textbf{Irrigation and Water Harvesting}
\begin{itemize}
    \item \textbf{Irrigation}: Establish small-scale drip or bucket irrigation systems to efficiently use water for crops.
    \item \textbf{Water Harvesting}: Construct water pans, tanks, or install gutters to collect rainwater for use during dry spells.
    \item Explore rock catchment systems or sand dams, which are commonly used in the region to retain water.
\end{itemize}

\item \textbf{Soil Health Management and Conservation}
\begin{itemize}
    \item Use manure, compost, and fertilizers to enrich soil and improve water retention.
    \item Employ soil conservation methods such as terracing, mulching, and planting cover crops to prevent erosion and nutrient depletion.
\end{itemize}

\item \textbf{Switching or Supplementing with Livestock Keeping}
\begin{itemize}
    \item Given the arid conditions, consider increasing reliance on livestock farming as an alternative to crop farming.
    \item Rear drought-tolerant breeds of goats and sheep, which are more suitable for Kitui's conditions.
\end{itemize}

\item \textbf{Mixed Cropping and Crop Diversification}
\begin{itemize}
    \item Practice mixed cropping systems (e.g., combining cereals with legumes) to maximize outputs and reduce the risk of complete crop failure.
    \item Diversify farm production by integrating high-value crops or vegetables into farming cycles.
\end{itemize}

\item \textbf{Community and Cooperative Participation}
\begin{itemize}
    \item Join local weather monitoring and farmer groups to get timely weather updates and share resources.
    \item Engage with agricultural extension officers for technical support and access to drought-resilient seeds.
\end{itemize}

\item \textbf{Switch to Water-Efficient Practices in Farming}
\begin{itemize}
    \item For a long-term strategy, lessons like permaculture or agroforestry can be implemented to improve resilience.
\end{itemize}
\end{enumerate}

By integrating these strategies into farming activities, farmers in Kitui, Kenya, can enhance their adaptability to climate variability, reduce crop failure risks, and improve agricultural resilience over time. If more specific issues arise (e.g., sudden rainfall, locust invasions, etc.), tailored solutions can be provided.

\end{document}